\documentclass[10pt,twocolumn,letterpaper]{article}
\usepackage{times}
\usepackage{graphicx}
\usepackage{amsmath}
\usepackage{amssymb}
\usepackage{amsthm}
\usepackage{multirow}
\usepackage{tabularx}
\usepackage{xcolor}
\usepackage{authblk}
\newcommand{\y}{\mathbf{y}}

\newcommand{\mbf}[1]{\mathbf{#1}}
\newcommand{\cut}[1]{}

\theoremstyle{definition}

\usepackage[pagebackref=true,breaklinks=true,colorlinks,bookmarks=false]{hyperref}

\begin{document}
\title{Integrating Propositional and Relational Label Side Information for Hierarchical Zero-Shot Image Classification}

\author[1]{Colin Samplawski}
\author[2]{Heesung Kwon}
\author[1]{Erik Learned-Miller}
\author[1]{Benjamin M. Marlin}
\affil[1]{University of Massachusetts Amherst}
\affil[2]{Army Research Laboratory}

\date{}

\maketitle
\begin{abstract}
Zero-shot learning (ZSL) is one of the most extreme forms of learning from scarce labeled data. It enables predicting that images belong to classes for which no labeled training instances are available.
In this paper, we present a new ZSL framework that leverages both label attribute side information and a semantic label hierarchy. We present two methods, lifted zero-shot prediction and a custom conditional random field (CRF) model, that integrate both forms of side information. We propose benchmark tasks for this framework that focus on making predictions across a range of semantic levels. We show that lifted zero-shot prediction can dramatically outperform baseline methods when making predictions within specified semantic levels, and that the probability distribution provided by the CRF model can be leveraged to yield further performance improvements when making unconstrained predictions over the hierarchy. 
\end{abstract}

\section{Introduction}
The zero-shot learning (ZSL) framework is one of the most extreme forms of learning from  scarce labeled data. It enables predicting that images belong to classes for which no labeled training instances are available. Standard ZSL methods accomplish this by introducing a source of propositional side information about the labels in the form of label attribute vectors. However, when ZSL methods are applied to data sets like ImageNet to simulate large numbers of classes with no labeled training instances, the resulting prediction accuracy can be quite low (between 1\% and 20\% depending on the precise evaluation scenario \cite{Frome2013,Chao2016,Xian2017,Wang2018}). 

We propose a new zero-shot learning framework that we refer to as \textit{hierarchical zero-shot learning} or \textit{HZSL}  to address some of the limitations of the standard ZSL framework. Specifically, hierarchical zero-shot learning integrates two distinct forms of side information: standard label attribute vectors and a semantic hierarchy over the class labels expressing ``is-a" relationships between classes. 

The semantic hierarchy can be used to enable prediction at different semantic levels, exposing the inherent trade-off between the level at which semantic predictions are made and the potential for making prediction errors. It also enables the definition of flexible prediction utility functions over the hierarchical structure to reflect preferences for making different accuracy-specificity trade-offs. For example, when a model is uncertain about what fine-grained class an object belongs to, we may prefer for it to issue a correct higher-level prediction than an incorrect fine-grained prediction. 

We propose two distinct classification methods within the HZSL framework. First, we propose ``lifted'' zero-shot prediction. This approach computes predictions using a standard ZSL method, and projects them into higher-level nodes in the hierarchy. We show that lifted zero-shot prediction can substantially improve prediction accuracy compared to applying standard ZSL methods to higher-level semantic categories.  Second, we develop a novel model within the conditional random field (CRF) family that enables a deep integration of labeled images, label attribute vectors, and the semantic label hierarchy. We show that this approach leads to further improvements in prediction accuracy. We also show that HZSL methods can cover the case of encountering test-time images whose true class (and thus label attribute vector) are completely unknown at training time.  

The primary  contributions of this work are as follows: (1) we formalize the HZSL framework for image classification, (2) we propose two distinct approaches to learning and prediction within the HZSL framework, (3) we develop a benchmark label hierarchy and several benchmark HZSL prediction tasks and performance measures, and (4) we conduct an extensive experimental evaluation comparing multiple approaches on multiple tasks.

We begin by reviewing related work in Section~\ref{sec:background}. We present the proposed framework in Section~\ref{sec:framework}, along with the 
lifted zero-shot prediction approach and the proposed CRF model. In Section~\ref{sec:benchmarks} we present the benchmark label hierarchy, benchmark learning tasks, and evaluation procedures. In Section~\ref{sec:experiments}, we present and discuss results.

\section{Background and Related Work}
\label{sec:background}

In this section, we begin by reviewing the formal definition of the standard image classification problem. We next define the standard zero-shot image classification problem and review prior approaches within this framework. Finally, we briefly review the relationship to open set classification.

\subsection{The Standard Image Classification Problem}

In the standard image classification problem, the input space $\mathcal{X}$ corresponds to a space of images, while the  space of labels $\mathcal{Y}$ corresponds to a set of mutually exclusive categories. The label $y$ associated with a given input $\mbf{x}$ typically represents the category of the most prominent object in $\mbf{x}$. In the standard image classification problem, a classifier is learned using only a data set $\mathcal{D}_{tr}=\{(\mbf{x}_n,y_n) \ | \  n=1,...,N \}$ of image-label pairs.
Further, it is assumed (typically implicitly) that the label set $\mathcal{Y}_{tr}$ represented in the training data set $\mathcal{D}_{tr}$  is exhaustive, corresponding to the assertion that $\mathcal{Y}=\mathcal{Y}_{tr}$. 
The formal definition of a standard classifier learning algorithm $L_{S}$ is then a function that maps a training data set $\mathcal{D}_{tr}$ to a classification function of the form $f: \mathcal{X}\rightarrow \mathcal{Y}_{tr}$.

However, when classifiers are deployed in real-world settings, there is a possibility of encountering images whose true labels lie outside of the set $\mathcal{Y}_{tr}$. We next turn to zero-shot learning methods, which aim to address this problem. 

\cut{Since the only information available during learning in this problem are the images and labels in the training data set $\mathcal{D}_{tr}$, this problem cannot be solved simply by expanding the label set $\mathcal{Y}_{tr}$ to $\mathcal{Y}$ during learning, even if the set of all possible labels is known. This is due to the fact that all standard learning algorithms  will drive the likelihood of predicting a class from the set $\mathcal{Y} \setminus \mathcal{Y}_{tr}$ to zero since $\mathcal{D}_{tr}$ contains no images with labels from that set. }

\subsection{The Zero-Shot Classification Problem}
\label{sec:background:zsl}

In the standard ZSL framework, learning methods again have access to a training data set $\mathcal{D}_{tr}$; however, they also have access to propositional side information in the form of real-valued class attribute vectors $\mbf{a}_y$. The key idea is that these class attribute vectors are provided for an expanded set of classes $\mathcal{Y}_{zs}$ containing $\mathcal{Y}_{tr}$ as well as other classes. The side information
is thus defined as $\mathcal{A} =\{\mbf{a}_y \ | \ y\in \mathcal{Y}_{zs}\}$. The inclusion of this side information allows for learning about classes in the set $\mathcal{Y}_{zs}~\setminus~\mathcal{Y}_{tr}$ for which training images are not available, but class attribute vectors are available. The formal definition of a ZSL learning algorithm $L_{ZS}$ is a function that maps a data set $\mathcal{D}_{tr}$ and the propositional side information $\mathcal{A}$ to a classification function of the form $f: \mathcal{X}\rightarrow \mathcal{Y}_{zs}$.

In early work on ZSL, data sets were created that included hand-curated attribute vectors $\mbf{a}_y$ for each class \cite{farhadi2009describing,lampert2009learning,lampert2014attribute}. The process of creating these attribute vectors is itself quite laborious, so recent work on this topic has focused on the use of attribute vectors learned using side data sets. A typical approach is to use a side corpus of text documents to learn unsupervised word vector embeddings for the word (or phrase) that forms the label for each class. 

Given a set of attribute vectors for each class in $\mathcal{Y}_{zs}$, a common approach to the zero-shot classification problem is compatibility learning. The compatibility learning problem attempts to learn a compatibility function 
$S: \mathcal{X} \times \mathcal{A} \rightarrow \mathbb{R}$ that maps images and label attribute vectors to real-valued scores. This approach can predict that an image belongs to a class from $\mathcal{Y}_s~\setminus~\mathcal{Y}_{tr}$ for which no images are seen during training if that class results in the maximum similarity.

For example, the deep visual-semantic embedding (DeViSE) model \cite{Frome2013} learns a linear compatibility function between image feature vectors and label attribute vectors $\mbf{a}_y$.
\cut{
\begin{equation}
C(\text{image } \mbf{x}, \text{label } y) = \mbf{a}_y^T \mbf{W} \phi(\mbf{x})
\end{equation}
where $\mbf{W}$ is a matrix of parameters. 
}
The resulting compatibility scores are then used to optimize a multiclass hinge loss objective. Intuitively, this can be thought of as learning a linear mapping from the image space into the label space and learning a compatibility function there. Linear compatibility learning is commonly used for ZSL \cite{Akata2013,Frome2013,Romera-Paredes2015,akata2015evaluation,akata2016label}, however work exists on non-linear compatibility learning as well \cite{Socher2013,xian2016latent,Zhang2017}. 

A related set of approaches eliminate explicit compatibility learning in favor of representing unseen classes as a weighted combination of training classes label embeddings \cite{Norouzi2013,zhang2015,changpinyo2016synthesized}.  An example is the convex combination of semantic embeddings (ConSE) method \cite{Norouzi2013}. This approach first applies a standard image classifier to an image, obtaining label probabilities $P(Y=y| \mbf{X}=\mbf{x})$ for all labels $y\in\mathcal{Y}_{tr}$. The image $\mbf{x}$ is then embedded using a convex combination of the vectors $\mbf{a}_y$ for the classes $y\in\mathcal{Y}_{tr}$ weighted by these probabilities. This embedding is used in a nearest neighbor search over the set of all test embeddings. The test embedding with the highest cosine similarity is returned as the prediction for $\mbf{x}$. 

The recent work of Wang et al \cite{Wang2018} performs ZSL using graph convolutional networks (GCNs) \cite{kipf2016semi}.
To use GCNs for ZSL, the required graph is generated from the ImageNet hierarchy. This model was able to show impressive improvements over previous baselines on zero-shot benchmarks.
\cut{After a series of layers, a $k$-dimensional representation is generated for each node in the graph, where $k$ is the number of features in the image representation. The parameters are then learned by taking the mean squared error between these outputs and ground truth images for those nodes.}

Most recently, approaches to ZSL using generative adversarial networks \cite{goodfellow2014generative} have been considered \cite{Zhu_2018_CVPR,Chen_2018_CVPR,Verma_2018_CVPR,Xian_2018_CVPR}. These models are able to generate features for classes without training images using side information such as textual descriptions. Once features for unknown classes have been generated, a standard classifier can be used to discriminate between classes. These approaches have primarily been applied to small, more focused datasets such as Animals with Attributes \cite{lampert2014attribute} or Caltech-UCSD Birds 200 \cite{WelinderEtal2010}. There has been limited application of these models on large scale datasets such as ImageNet, which we use in this work.

The ability to define models that can make predictions outside of the set of labels seen within the training data set is an important advance relative to the standard classification problem setting. However, the zero-shot framework is still limiting in the sense that the label set $\mathcal{Y}_{zs}$ used in the ZSL setting must contain mutually exclusive labels. In addition, it is still possible to encounter images that belong to classes outside of $\mathcal{Y}_{zs}$ at deployment time. Further, the application of zero-shot methods to scenarios where $|\mathcal{Y}_{zs} \setminus \mathcal{Y}_{tr}|$ is large typically yields poor performance. Indeed, state-of-the art methods typically obtain accuracy rates between 1\% and 20\% on ImageNet depending  on the exact evaluation procedure \cite{Frome2013,Chao2016,Xian2017,Wang2018}.  

\subsection{The Open Set Classification Problem}
Open set classification is a different approach to the problem of encountering previously unseen classes at test time based on attempting to predict when classes seen at test time lie outside of $\mathcal{Y}_{tr}$.\cut{To accomplish this, the open set problem can be thought of as augmenting the training classes $\mathcal{Y}_{tr}$ with a single additional class label $\emptyset$ that is used to indicate that an input belongs to a class that is outside of $\mathcal{Y}_{tr}$. The complete set of classes considered in the open set framework is thus $\mathcal{Y}_{os} = \mathcal{Y}_{tr} \cup \{\emptyset\}$.} The advantage of this framework is that it makes no assumptions about the number or identity of the unknown classes. However, a disadvantage is that it makes no attempt to use available side information known about labels for which there are no training images.

Previous approaches to open set classification consider minimizing the ``open set risk" of a classifier as well as the traditional empirical risk. Most commonly this is accomplished using margin based methods which include a margin to separate the known space from the open space \cite{Scheirer_2013_TPAMI, Bendale2015, Rudd2018}. 

In this work, we will consider instances of classes that are completely unknown (outside of $\mathcal{Y}_{zs}$) at training time, but we will do so using an orthogonal approach based on mapping these instances into higher levels of a known semantic hierarchy.

\cut{
\subsection{ImageNet Background}
We provide a brief summary of the various subsets of ImageNet  \cite{imagenet}  that are commonly used in benchmarks. The most common set of classes is the 1,000 ImageNet Large Scale Visual Recognition Challenge (ILSVRC) classes consisting of about 1.3 million images. Classification benchmarks are so common on this set that is it often referred to simply as ImageNet classification. However, the full ImageNet hierarchy is made up of $\sim$32,000 classes, $\sim$21,000 of which have images. This hierarchy presents a graph of ``is-a'' relations between labels. For example, the label ``car" represents the category of all possible cars with the parent label ``wheeled vehicle" representing all possible wheeled vehicles, implying the relation is-a(car, wheeled vehicle).

For zero-shot  benchmarks it is common to draw the set of unseen classes from the set of $\sim$20,000 classes that don't include the 1,000 ILSVRC classes. Popular test sets of this form are the ``2-hops'' and ``3-hops'' test sets \cite{Frome2013}. These sets consist of classes that are within 2 or 3 tree hops from one of the 1,000 ILSVRC classes in the ImageNet hierarchy, with 1,549 and  7,861 classes respectively. Note that 2-hops is a subset of 3-hops.
}

\section{The HZSL Framework }
\label{sec:framework}

In this section, we describe the proposed hierarchical zero-shot learning (HZSL) framework along with two learning and prediction approaches for this framework: lifted zero-shot learning and a conditional random field model.

\subsection{The HZSL Learning Framework}
The standard zero-shot image classification framework leverages propositional information in the form of a set of label attribute vectors $\mathcal{A}$ for an expanded label space $\mathcal{Y}_{zs}$ as described in Section \ref{sec:background:zsl}. The proposed HZSL framework leverages the same propositional side information used in standard ZSL, but relaxes the requirement that the label set $\mathcal{Y}_{zs}$ contain mutually exclusive labels. HZSL simultaneously introduces an additional source of relational side information about the labels in the form of a semantic label hierarchy $\mathcal{H}$. 

The semantic label hierarchy $\mathcal{H}$ is defined as a set of ordered tuples of class labels where if $(y,y')\in \mathcal{H}$, then  $y'$ is said to be the parent of $y$. Semantically, we interpret the $(y,y')\in \mathcal{H}$ relation to assert that the category represented by $y$ is a subset of the category represented by $y'$. We further assert that each class $y$ has exactly one parent in $\mathcal{H}$, except for a universal class $\Omega$, which contains all other possible categories and has no parent.

Together, these requirements result in a tree structure over the space of labels where the edges in the tree correspond to ``is-a" relationships between labels. We let $\mathcal{Y}_{hzs}$ represent the set of all labels within the hierarchy $\mathcal{H}$, which will typically include the labels $\mathcal{Y}_{zs}$. The formal definition of an HZSL learning algorithm $L_{HZS}$ is then a function that maps a data set $\mathcal{D}_{tr}$ and the tuple of propositional and relational side information $(\mathcal{A},\mathcal{H})$ to a classification function of the form $f: \mathcal{X}\rightarrow 
\mathcal{Y}_{hzs}$. The semantic relationships that underlie $\mathcal{H}$ imply that if we predict that $f(\mbf{x})=y$, we have also predicted that $\mbf{x}$ is an instance of each class $y'$ that is an ancestor of $y$ in $\mathcal{H}$. \cut{As with standard zero-shot learning, the resulting HZSL classifier $f$ can make meaningful predictions for classes that lie outside of $\mathcal{Y}_{tr}$ due to the use of side information regarding the expanded label set $\mathcal{Y}_{hzs}$.}

Finally, we note that the use of such a semantic hierarchy with a universal root node $\Omega$ can side-step the problem of encountering an image with a label $y\not\in\mathcal{Y}_{hzs}$. Specifically, a class $y$ that does not exist in $\mathcal{Y}_{hzs}$ can still be predicted to belong to a class $y'$ that is its ancestor in $\mathcal{H}$.  Such a prediction may be more semantically meaningful in some applications than either predicting that an instance of class $y$ is out-of-set, or being forced to make a prediction among a set of known, mutually exclusive fine-grained classes that does not include $y$. We will evaluate the ability of HZSL methods to make predictions under this scenario, as well as under scenarios where the test time classes are drawn from $\mathcal{Y}_{tr}$ and $\mathcal{Y}_{hzs}$. We next turn to the development of learning and prediction methods for the HZSL framework.

\subsection{Lifted ZSL Learning and Prediction}
The first approach we describe to learning in the HZSL framework is lifted zero-shot learning. Lifted ZSL requires access to a standard ZSL algorithm $L_{ZS}$ as described in Section \ref{sec:background:zsl}. Given a data set $\mathcal{D}_{tr}$ and a set of label attribute vectors $\mathcal{A}$, we apply $L_{ZS}$ to learn a zero-shot classification function $f$.

To enable hierarchical prediction, we project the prediction $f(\mbf{x})$ into a desired level $l$ of $\mathcal{H}$ simply by identifying the ancestor of $f(\mbf{x})$ at level $l$ of $\mathcal{H}$. To do so, we define the function $\mbox{ancestor}(y,l)$ to return the ancestor of node $y$ at level $l$ of $\mathcal{H}$. The final lifted prediction function is given by $g(\mbf{x})=\mbox{ancestor}(f(\mbf{x}),l)$. In this approach, $l$ is a parameter of the lifted prediction method and can be varied to produce predictions at different semantic levels.

A limitation of lifted ZSL is that it can not provide a probability distribution over the full label hierarchy, which is a useful capability for supporting decision making under uncertainty. We now turn to the development of the proposed conditional random field model, which can provide a probability distribution over the full label hierarchy.

\subsection{The HZSL CRF Model}
In this section, we introduce an HZSL approach based on a conditional random field (CRF) model \cite{Lafferty2001} whose structure is given by the label hierarchy $\mathcal{H}$. This model has the advantage that it can use features output by multiple standard ZSL methods and image classification models, while defining a conditional probability distribution over $\mathcal{H}$ given an input image.

\noindent\noindent$\bullet$ \textbf{Random Variables:} A CRF is a probabilistic model defined in terms of two sets of random variables: label variables and feature variables. We define the set of label variables for the CRF to be the set of binary random variables $\mbf{Y}=\{Y_c \ | \ c \in \mathcal{Y}_{hzs}\}$. We let $\mbf{y}$ represent a realization of these label random variables. The model is conditioned on an input image $\mbf{x}$ that is an instance of an image random variable $\mbf{X}$.

\noindent\noindent$\bullet$ \textbf{Conditional Probability Distribution:} The conditional probability distribution induced by a CRF over the set of label variables $\mbf{y}$ given an image $\mbf{x}$ is specified through an energy function $E_{\theta}$ as shown below.

\begin{equation}
p_{\theta}(\mbf{Y}=\y | \mbf{X}=\mbf{x}) = \frac{ \exp (-E_{\theta}\left(\y, \mbf{x})\right)}
{\sum_{\y'} \exp \left(-E_{\theta}(\y', \mbf{x})\right)}
\end{equation}

\noindent The denominator of the above probability is referred to as the the \textit{partition function}. In a general CRF model, it is a sum over all $2^{|\mbf{y}|}$ possible configurations of the binary label vector $\mbf{y}$. The energy function $E_{\theta}$ that defines the model has parameters $\theta=[\mbf{w},b]$  and is a weighted combination of a set of $|\Phi|$  vector-valued feature functions $\phi(\mbf{x}, \mbf{y})$ as shown below. The core of the definition of a CRF model is thus the specification of the feature functions, which we describe next. 
\begin{equation}
E_{\theta}(\y, \mbf{x}) = \sum_{\phi\in\Phi} \mbf{w}_{\phi}^T \phi(\mbf{x}, \mbf{y})  + b
\end{equation}

\noindent$\bullet$ \textbf{Feature Functions:} In this model, we use three sets of feature functions for each class $c$ to capture information about the compatibility between class $c$ and the input image $\mbf{x}$.

We first introduce a basic linear feature $\phi^L_c(\mbf{x}, \mbf{y})= y_c \cdot 
\left( \mbf{W} h(\mbf{x}) \right)_c$, with a set of additional weights $\mbf{W}$. 
In this work, we choose $h(\mbf{x})$ to be the final hidden activations over $\mathcal{Y}_{tr}$  output by a standard CNN classifier trained on ImageNet. This representation preserves the information contained in a standard CNN about the classes in $\mathcal{Y}_{tr}$ while enabling other classes to be recognized via characteristic activation patterns.

Our second feature $\phi_{c}^D$ is defined via a compatibility function $S_D(\mbf{x}, c)$ between images and labels that is a non-linear extension of the DeViSE model introduced for standard ZSL. The definition is $\phi_{c}^D(\mbf{x}, \mbf{y}) = y_c\cdot \log S_D(\mbf{x}, c)$. The specific compatibility function that we use is given by a two-layer neural network:
\begin{equation}
S_D(\mbf{x}, c) = \frac{\exp\left(a_{c}^T g(\mbf{W}_2\cdot g(\mbf{W}_1 h(\mbf{x}) ))\right) }
{\displaystyle \sum_{c'\in \mathcal{Y}_{zsl}}\exp\left( a_{c'}^T g(\mbf{W}_2\cdot g(\mbf{W}_1 h(\mbf{x}) )) \right)} 
\end{equation}
This model leverages the label attribute vectors $\mbf{a}_c$ as well as the same image representation $h(\mbf{x})$ described above. The function $g$ is a non-linearity. This feature function provides information available via standard zero-shot compatibility approaches to the CRF model.

Next, we add a feature function $\phi_{c}^C$ inspired by the ConSE model developed for standard ZSL. This feature provides an additional source of label embedding similarity information to the CRF. We again use the representation $h(\mbf{x})$ defined above, but this time to generate a convex combination of training class label attribute vectors $\epsilon(\mbf{x})$ as a representation for $\mbf{x}$ in the label space. Using this new label vector, we calculate the cosine similarity with each label attribute vector in $\mathcal{H}$. Let $\{\pi_i\}^m_{i=1}$ be the probabilities for the $m$ most likely training classes as given by the softmax output layer of an image classifier for $\mbf{x}$ and $\{\mbf{a}_i\}^m_{i=1}$ be the label attribute vectors for those classes. The embedding of $\mbf{x}$, $\epsilon(\mbf{x})$, is then
    $\epsilon(\mbf{x}) = \frac{1}{Z} \sum_{j=1}^m \pi_i \mbf{a}_i$
where $Z$ is a normalizing constant given by $Z=\sum_{j=1}^m \pi_i$. 
The feature function is then $\phi_{c}^C(\mbf{x}, \mbf{y}) = y_c \cdot \text{cosinesim}(\mbf{a}_c, \epsilon(\mbf{x}))$.

\begin{figure}[t]
	\centering
	\includegraphics[width=.9\linewidth]{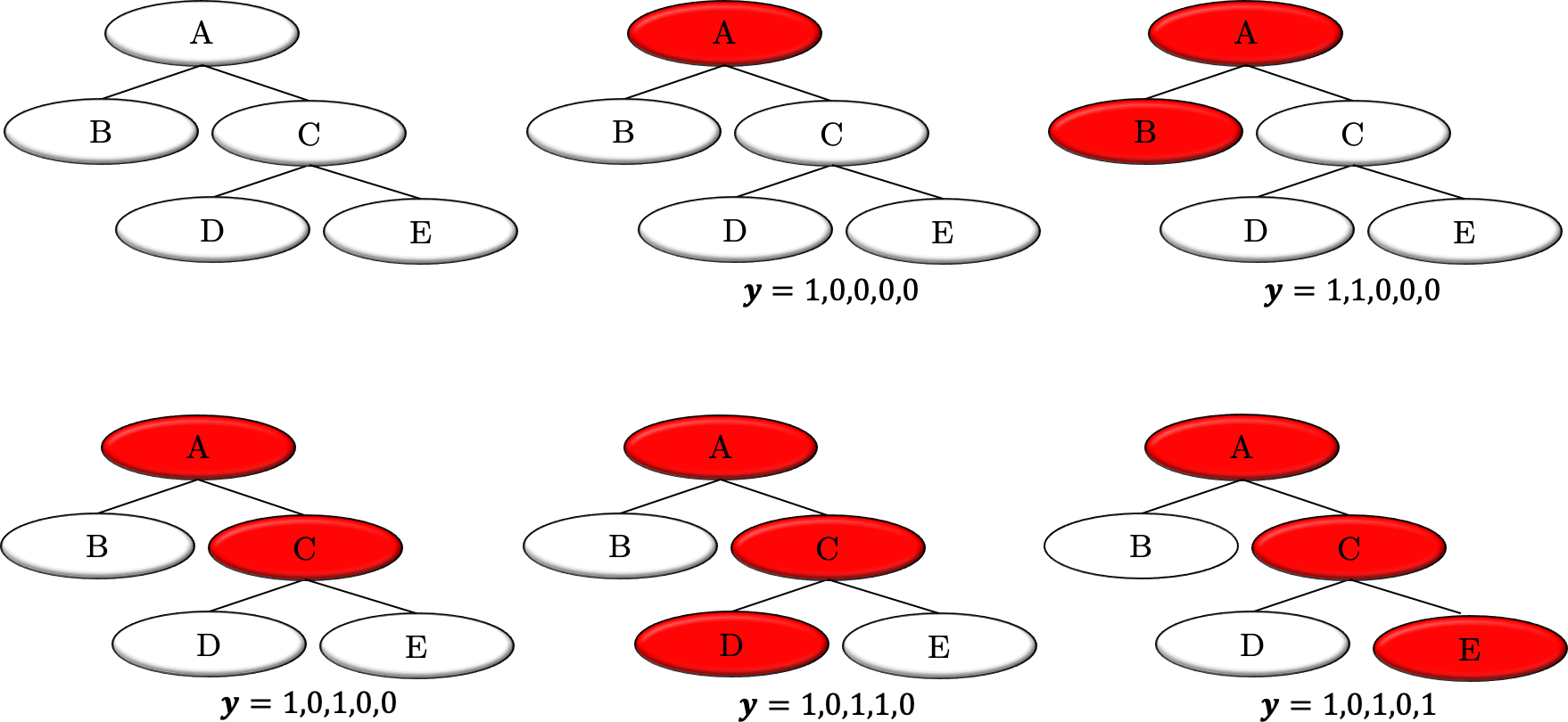}
	\caption{A simple hierarchy with five possible paths. Here we display those paths and the corresponding binary label vector $\mbf{y}$. Our CRF model outputs a probability distribution over them.}
	\label{fig:possible_paths}
\end{figure}

A final feature function $\phi^G(\mbf{x},\mbf{y})$ is required to ensure that joint label configurations respect the nesting required by the ``is-a" semantics of the label hierarchy. In particular, when an image is of class $c$, we must have that $Y_c=1$, $Y_{c'}=1$ for all ancestors $c'$ of $c$, and $Y_i=0$ for all other labels $i$. All other configurations of the label variables have probability $0$. The set of allowable configurations of the labels thus corresponds exactly to the set of shortest paths from the root node to each node in $\mathcal{H}$. Figure \ref{fig:possible_paths} illustrates this set of paths for a simple five class hierarchy. 

We define $\phi^G(\mbf{x},\mbf{y})=0$ if the label configuration in  $\mbf{y}$ corresponds to a valid path in $\mathcal{H}$ and $-\infty$ otherwise. Fortunately, the inclusion of this higher-order factor actually makes learning and prediction with the CRF model easier because including it is equivalent to simply normalizing the CRF over the set of valid paths in $\mathcal{H}$, and this set of paths has size exactly equal to $|\mathcal{Y}_{hzs}|$.

\noindent$\bullet$ \textbf{Learning and Prediction:} For a training set of $N$ labeled images $\mathcal{D}_{tr}$, we learn the model parameters $\theta$ by minimizing the average negative conditional log likelihood under the CRF model \cite{Lafferty2001}:
\begin{equation}
\mathcal{L}(\theta|\mathcal{D}_{tr}) = -\frac{1}{N} \sum_{i=1}^{N} \log p_{\theta}(\y_n| \mbf{x}_n).
\end{equation}
All the components of this objective function are differentiable, so we can optimize the model parameters via standard gradient-based methods. As noted above, the CRF partition function is exactly computable in time $O(|\mathcal{Y}_{hzs}|)$, which is easily tractable.

To make predictions in the learned model, we begin by computing the probability of every valid path in the hierarchy. We can then select the highest probability path in the model and output the last label on that path as the model's prediction. If we are interested in restricting the predictions to a given level in the hierarchy, we can do so by selecting the most  likely label at that level. Since each higher level label represents a subtree of the hierarchy under this scenario, we can compute the probability mass associated with a subtree simply by summing the probability of all paths passing through that node. Finally, given a prediction utility function $U$, we can select as our prediction the node in the hierarchy with the highest expected utility under the model. 

\subsection{Implementation Details} In the experiments that follow, we use a pre-trained ResNet101 \cite{He2016} model trained to classify the 1,000 ImageNet classes. For the image feature function $h(\mbf{x})$, we extract the output of the final pooling layer of this model, resulting in a 2048-dimensional feature vector. 

For the label attribute vectors $\mbf{a}_c$ we train a Subword Information Skip Gram model \cite{Bojanowski2017} on all of English Wikipedia using the fastText library.\footnote{\url{https://fasttext.cc}} This model is an extension of the skip-gram with a negative sampling word2vec model \cite{Mikolov2013a,Mikolov2013b} which incorporates sub-word information in the form of $n$-grams. This allows for the generation of embeddings for words that do not appear in the training data. This is useful for our purposes, since a handful of ImageNet labels fall outside of the pre-trained GloVe  embedding set \cite{Pennington2014}, which is commonly used in the ZSL literature. Before training the model, we preprocessed the text and replaced all multiword labels with a single token, e.g.\ ``sea lion'' becomes ``sea\_lion''. This leads to potentially richer presentations for multiword labels compared to simple techniques such as vector averaging. 

We pre-train the DeViSE-like compatibility model on the 1,000 class ImageNet training set in the same way as previous ZSL work. We parameterize the compatibility function with a two-layer feed forward neural network with ReLU activation functions. When training the CRF, we backpropagate through the CRF and compatibility model parameters. The weights which generate the ResNet101 image representations and classification probabilities remain fixed throughout training in our experiments, although in principle they could also be fine-tuned.
\section{Benchmarks for HZSL}
\label{sec:benchmarks}
In this section, we present the proposed benchmark data sets, label hierarchy and prediction tasks.

\subsection{ImageNet Background}
We begin by providing a brief summary of the various subsets of ImageNet  \cite{imagenet}  that are commonly used in classification benchmarks. The most common set of classes is the 1,000 ImageNet Large Scale Visual Recognition Challenge (ILSVRC) classes consisting of about 1.3 million images. Classification benchmarks are so common on this set that is it often referred to simply as \textit{ImageNet classification}. However, the full ImageNet graph is made up of about 32,000 classes, about 21,000 of which have images. This graph represents ``is-a'' relations between labels.

For zero-shot  benchmarks, it is common to base the set of unseen classes on the set of 20,000 classes with images that is disjoint from the 1,000 ILSVRC classes. Popular test sets of this form are the ``2-hops'' and ``3-hops'' test sets \cite{Frome2013}. These sets consist of classes that are within 2 or 3 tree hops from one of the 1,000 ILSVRC classes in the ImageNet graph, with 1,549 and  7,861 classes respectively. Note that the 2-hops classes are a subset of the 3-hops classes.

\subsection{Benchmark Data Set  and Semantic Hierarchy}

We propose a benchmark semantic hierarchy for ImageNet that includes the 1,000 ILSVRC classes. We start with the full WordNet \cite{wordnet} graph of nouns, of which ImageNet is a subgraph. We then iteratively remove leaf nodes from the graph until the only remaining leaves are either in the 1,000 ILSVRC classes or the set of nodes reachable from the 1,000 ILSVRC classes  within two hops (the 2-hops set). This results in a graph which contains non-tree edges (i.e.\ some nodes have more than one parent). We enforce a  tree structure by running Chu-Liu/Edmonds' algorithm \cite{Edmonds1967}, which finds an optimal spanning  tree given a general graph as input. This results in a tree with 2,831 nodes, which we use as the benchmark label hierarchy $\mathcal{H}$ and known label set $\mathcal{Y}_{hzs}$ in our evaluations.

In the resulting tree, approximately half of the leaf nodes come from the 2-hops set. We treat these classes as having known labels in the sense that at training time their location in the hierarchy and their label attribute vectors are known; however, no images of these classes are seen at training time (i.e.\ these nodes are in $\mathcal{Y}_{hzs} \setminus \mathcal{Y}_{tr}$). The rest of the tree is considered completely known (i.e.\ the remaining nodes are in $\mathcal{Y}_{tr}$). As we will see, we can also accommodate test images from classes that are not in $\mathcal{Y}_{hzs}$ at all.

\subsection{Benchmark HZSL Tasks}
\label{sec:benchmarks:tasks}

In this section, we introduce three benchmark tasks for HZSL that differ in terms of constraints on where in the semantic hierarchy predictions are made and what classes are included in the test sets. For all tasks, models are learned using the ILSVRC training set and the benchmark hierarchy described in the previous section.

\noindent$\bullet$ \textbf{Fine-Grained Classification:} In this task, we restrict the predictions to the fine-grained classes in $\mathcal{H}$. This task has two variants. We can draw the test instances used from the set of training classes $\mathcal{Y}_{tr}$ or from the 2-hops set. When making prediction on instances from the training classes, this is the standard ImageNet classification task. When making predictions on instances of the 2-hops classes, this is traditional zero-shot classification. We include this task to assess the performance of HZSL methods on standard tasks relative to standard models for those tasks.

\noindent$\bullet$ \textbf{Classification Within Higher Semantic Levels:} A motivating use case for HZSL methods is to provide more accurate predictions at higher levels in the semantic hierarchy. We therefore propose a family of tasks for this scenario. We begin by selecting 17 disjoint first-level classes that represent broad categories of common objects, such as ``mammal'', ``vehicle'', and ``person'', that cover the majority of the 1,000 ILSVRC classes. We generate a set of second-level categories by drawing from the direct children of the first-level categories. We do the same with the second-level categories to generate third-level categories. We end up with a set of 77 second-level categories and 134 third-level categories. Note that the lifted zero-shot models make their initial prediction $f(\mbf{x})$ over all the classes in $\mathcal{H}$ which fall under a higher-level class.

We propose three specific benchmarks for each semantic level by varying the sets of test instances considered. In addition to the choice of semantic level, we can also draw the test instances from different sets of classes. We consider this task using the training classes, the 2-hops classes, and the 3-hops classes. When using the 2-hops classes, we are performing higher-level zero-shot learning and when using the 3-hops classes we are performing a version of open set classification.

\noindent$\bullet$ \textbf{Free Hierarchical Classification:} In this set of tasks, we consider the ability of models to make free predictions across the complete semantic label hierarchy. Models can predict any label at any level. We again consider this task using the the 2-hops classes.

\noindent$\bullet$ \textbf{Data Set Statistics:} Table \ref{table:dataset_stats} provides a summary of the benchmark training and test sets described above. We note that not every class from the original test sets is included in our test sets. For the sake of evaluation, we only include classes which fall under one of the top-level categories described above. Furthermore, in the case of 3-hops, we consider classes that only fall into 3-hops and not 2-hops (i.e.\ the set 3-hops $\setminus$ 2-hops).
\begin{table}
	\centering
	\begin{tabular}{|l|r|r|}
		\hline
		Dataset & \# Classes & \# Images \\ \hline
		ILSVRC  for training & 1,000 & $\sim$1,300K \\
		ILSVRC  for testing & 657 & $\sim$35K\\
		2-hops  for testing & 820 & $\sim$600K \\
		3-hops  for testing & 2,837 & $\sim$2,000K\\ \hline
	\end{tabular}
	\caption{Training and test set statistics for benchmark tasks.}
	\label{table:dataset_stats}
\end{table}

\begin{figure*}[t]
	\centering
	\includegraphics[width=0.33\linewidth]{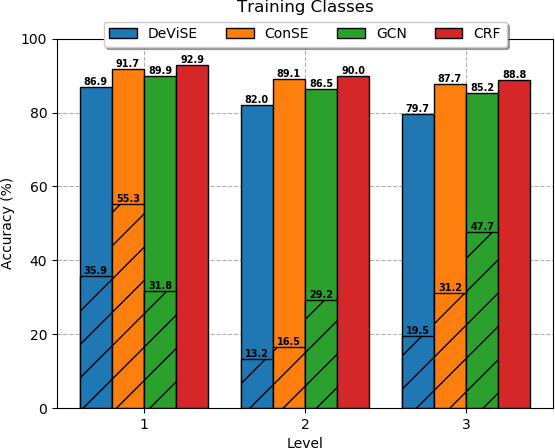}
	\includegraphics[width=0.33\linewidth]{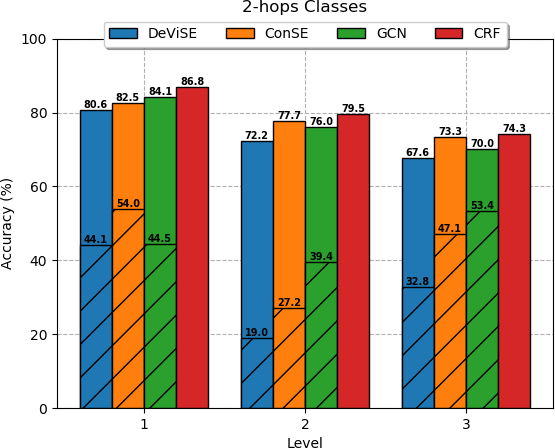}
	\includegraphics[width=0.33\linewidth]{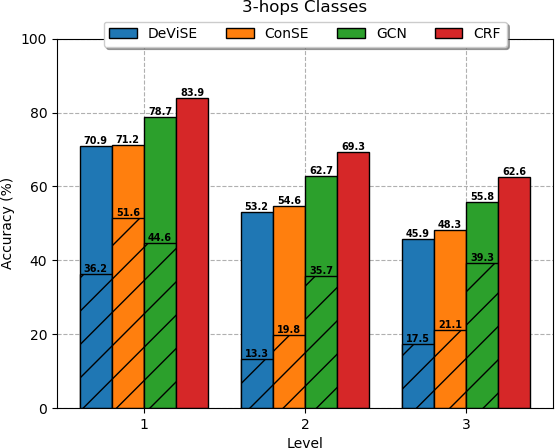}
	\caption{Classification within semantic level benchmark results. 
	The cross-hatched bars indicate the performance of non non-lifted version 
	of the baseline ZSL models.}
	\label{fig:category_prediction}
\end{figure*}

\begin{figure}[t]
	\centering
	\includegraphics[width=0.6\linewidth]{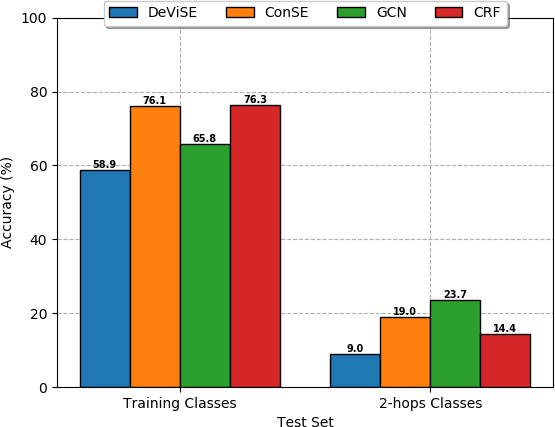}
	\caption{Fine-grained classification benchmark results.}
	\label{fig:finegrained}
\end{figure}

\section{Experiments and Results}
\label{sec:experiments}
In this section, we present experiments and results. We compare the proposed hierarchical zero-shot CRF model and the proposed lifted zero-shot prediction method using a number of base ZSL models from the literature including DeViSE \cite{Frome2013}, ConSE \cite{Norouzi2013}, and graph convolutional networks (GCNs) \cite{Wang2018}. We use the same image features and word embeddings for all models. For ConSE, we use $m=10$ for our experiments, since it was the best performing value in the original paper. The GCN model used the same 6-layer configuration as in the original paper. We ensured that all three models produced similar or better performance as published on standard ZSL benchmarks before using them in our evaluations.\\

\noindent$\bullet$ \textbf{Experiment 1: Fine-grained Classification} 
To begin, we consider the performance of each method on the fine-grained training classes benchmark. As described in Section \ref{sec:benchmarks:tasks}, this corresponds to standard ImageNet classification. The results are shown in in Figure \ref{fig:finegrained}. We include these results to illustrate how various methods perform on the set of completely known classes at test time. We see that the proposed CRF model and ConSE are able to preserve the majority of the performance provided by the underlying ResNet101 model (which has accuracy 76.4\%), while DeViSE and GCNs suffer significant performance drops. This is not desirable in situations where the training classes $\mathcal{Y}_{tr}$ are expected to occur at test time.

We next consider the performance of each method on the fine-grained 2-hops classes benchmark. As described in Section \ref{sec:benchmarks:tasks}, this corresponds to standard ImageNet zero-shot classification. The results are shown in in Figure \ref{fig:finegrained}. As in previous ZSL work, we see that accuracy is very low on this task relative to the standard classification task with GCNs showing the best performance at just below 24\% accuracy. As noted previously, our main focus is in mitigating this drop in accuracy by trading lower specificity of predictions for higher accuracy of predictions, which we explore in the following experiments.

\noindent$\bullet$ \textbf{Experiment 2: Classification Within Semantic Levels} In Figure \ref{fig:category_prediction}, we present the results of the classification within semantic levels benchmark. The three panels correspond to test instances drawn from the training classes (left), the 2-hops classes (middle) and the 3-hops classes (right). Within each panel, results are shown for lifted zero-shot prediction methods and the CRF. 

For comparison, we also show the results for a direct application of the zero-shot methods within the specified semantic level. For example, at semantic level 1, each model considers the 17 word embeddings for the categories ``mammal'', ``vehicle'', ``person'', etc. and outputs the most compatible category. Intuitively, we are using the baseline ZSL models to make a ZSL prediction over the set of word embeddings that make up the first-level categories. These baselines have no access to information about the label hierarchy. 

First, comparing to Figure \ref{fig:finegrained} to Figure \ref{fig:category_prediction}(middle), we can see that the direct application of zero-shot methods within higher semantic levels on the 2-hops test set (the cross-hatched bars) results in improvements of 10 to 35\% depending on semantic level. This is the direct result of trading decreased specificity of zero-shot predictions for improved accuracy. Next, comparing the direct application of zero-shot methods within higher semantic levels (cross-hatched bars) to lifted zero-shot prediction within semantic levels (solid bars) in Figure \ref{fig:category_prediction}, we see that the proposed lifting approach results in further highly significant performance improvements across all test sets and base ZSL methods. The improvements over the direct application of ZSL within semantic levels is between 15 and 70\% depending on test set and task. The large performance improvement obtained by lifted zero-shot prediction is explained by the fact that the direct ZSL models are basing their predictions on the structure of the word embedding space, and co-occurence word embedding methods offer no guarantees that hierarchical relationships between concepts are expressed in the embedding space. Lifted zero-shot prediction overcomes the limitations of standard label embeddings of higher-level categories by assessing similarity at the fine-grained level where these similarity computations are more meaningful, and then lifting the resulting predictions to the desired level.

Second, we see that the proposed CRF model achieves a further improvement over the lifted ZSL models, with a performance increase ranging from 1 to 7\% over the next best model. For perspective, correctly classifying an additional 7\% of the 3-hops images corresponds to correctly classifying an additional $\sim$150,000 images.

Finally, we note that all models see a noticeable drop in performance between the training classes, 2-hops and 3-hops versions of these benchmarks. This is not surprising since many objects in the 3-hops set are semantically further away from the training classes than the 2-hops classes, which results in harder test cases. We also note that, in general, the lifted ZSL methods see steeper drops in performance as the test set requires more fine grained predictions when compared to the CRF. This suggests that the way that the CRF makes use of the relational side information provides an additional benefit over the proposed lifted prediction methods.

\noindent$\bullet$ \textbf{Experiment 3: Free Hierarchical Classification}
\begin{table}
	\centering
	\resizebox{\columnwidth}{!}{
	\begin{tabular}{|c|c|c|c|c|}
		\hline
		Model & Accuracy & Path Length ($U_{PL}$)& Subtree Depth ($U_{SD}$)\\ \hline
		DeViSE & 0.012 & 0.727 &  0.395 \\
		ConSE & 0.003  & 0.766 & 0.523 \\
		GCN & \textbf{0.095} & 0.751 & 0.339 \\
		CRF & $\sim$0.0 & \textbf{0.811} & \textbf{0.537}\\ \hline
	\end{tabular}
	}
	\caption{Average utility generated by the CRF and baselines for the utility functions discussed. Results are given as  means over the per image utility for images from our subset of the 2-hops classes.}
	\label{table:rewards}
\end{table}
In the experiments discussed above, we measure a model's ability to make predictions at pre-specified levels of the hierarchy in terms of accuracy. However, it may be more desirable to allow greater flexibility when making predictions. In this experiment, we allow models to freely make predictions at any level of the hierarchy. 

The CRF model has the ability to make such predictions natively as it produces a properly normalized probability distribution over all classes in the hierarchy. However, lifted zero-shot methods can not be applied in this task since they require the specification of a semantic level. Instead, we again perform a direct application of zero-shot methods, but this time we compute compatibility scores for all nodes in $\mathcal{H}$ and select the node with the highest score.

Next, we note that standard prediction accuracy is unlikely to be sufficient to assess prediction quality in this setting since it does not differentiate between different kinds of prediction errors. Instead, we introduce a more general notion of the utility associated with a prediction. We define the function $U(y',y)$ to be the utility of predicting label $y'$ for an example with ground truth path $y$. 

Given access to a model that can induce a probability distribution over the label hierarchy $\mathcal{H}$, we can define the expected expected utility of predicting label $y'$ as shown below. A prediction can then be made based on the label that maximizes the expected utility. 

\begin{equation}
U(y') = \sum_{y\in\mathcal{Y}_{zhs}} P(Y=y | \mbf{x}) \cdot U(y',y)
\end{equation} 
The question then becomes selecting a utility function that matches the desired trade-offs for a given application. We discuss three examples below.

One refinement of prediction accuracy for a semantic hierarchy is to consider how close the predicted class is from the true class in terms of shortest path distance in the hierarchy. The corresponding utility $U_{PL}(y',y)$ is then one minus the shortest path distance from  $y$ to $y'$ normalized by the length of the longest path in the hierarchy.

Our motivating examples correspond to preferring to correctly predict an ancestor of $y$ than to making other types of prediction errors. We can express this preference using a utility that is $0$ if the predicted label $y'$ is not an ancestor of $y$. When $y'$ is an ancestor of $y$, we should have higher utility the closer that $y'$ is to $y$ in the hierarchy. Below, we define the subtree depth utility $U_{SD}(y',y)$ that is $0$ when $y'$ is not an ancestor of $y$ and otherwise is ratio of the depth of $y'$ divided by the depth of $y$ in the hierarchy. Intuitively,  $U_{SP}(y',y)$ decreases monotonically for classes $y'$ along the path from $y$ to the root node $\Omega$. 
\begin{equation}
U_{SD}(y',y)=
\begin{cases}
0 & y' \text{ is not an ancestor of } y\\
\frac{depth(y')}{depth(y)} & \text{else}
\end{cases}
\end{equation}
This utility therefore strongly enforces the idea that a correct prediction should at the very least be a super category of the true label.

Table \ref{table:rewards} shows accuracy as well as average utility for $U_{PL}$ and $U_{SD}$ using under our CRF model and the baseline models. Here accuracy is analogous to the task of generalized zero-shot prediction \cite{Chao2016}, where an unseen class must be predicted from a search space which includes known classes. Note that since the baseline models are not probabilistic models, we cannot make predictions that maximize expected utility, instead we calculate the utility of the highest scoring prediction directly output by each method. 

As we can see, the CRF obtains the lowest accuracy on this task. GCNs perform the best in terms of accuracy, but the level of accuracy achieved is still very low. When we consider the other two utility functions, we can see that the CRF in fact achieves the highest average utilities. The reason is that it tends to be more conservative when making predictions and thus while it misses predicting the exact class nearly all of the time, it is generally closer  than the other models ($U_{PL}$), and tends to also predict correct ancestors more often ($U_{SD}$) relative to the baseline methods.  By contrast, we see  that  GCNs perform the worst of all of the methods on $U_{SD}$, indicating that when it makes a mistake, the predicted class tends to not be an ancestor of the  true class. 

Finally, we note that the utility functions explored here are only intended to provide examples of utility functions for hierarchical prediction. The utility framework provides a highly flexible approach to fine-tuning the predictions from the CRF model and can use customized utilities to meet the needs of other application scenarios.
\section{Conclusion}
We have presented a novel extension of zero-shot learning that leverages both propositional and relational side information to improve prediction utility. We have presented two different approaches to the resulting learning problem, and described benchmark data sets and tasks. Our results show that the proposed methods outperform applications of baseline ZSL methods when making predictions at a range of semantic levels, with the proposed CRF model showing the best overall performance.

{\small
\bibliographystyle{ieeetr}
\bibliography{paper}

\begin{thebibliography}{10}

\bibitem{Frome2013}
A.~Frome, G.~S. Corrado, J.~Shlens, S.~Bengio, J.~Dean, M.~Ranzato, and
  T.~Mikolov, ``Devise: A deep visual-semantic embedding model,'' in {\em
  NIPS}, pp.~2121--2129, 2013.

\bibitem{Chao2016}
W.-L. Chao, S.~Changpinyo, B.~Gong, and F.~Sha, ``An empirical study and
  analysis of generalized zero-shot learning for object recognition in the
  wild,'' in {\em ECCV}, pp.~52--68, 2016.

\bibitem{Xian2017}
Y.~Xian, B.~Schiele, and Z.~Akata, ``Zero-shot learning - the good, the bad and
  the ugly,'' in {\em CVPR}, pp.~3077--3086, 2017.

\bibitem{Wang2018}
X.~Wang, Y.~Ye, and A.~Gupta, ``Zero-shot recognition via semantic embeddings
  and knowledge graphs,'' in {\em CVPR}, pp.~6857--6866, 2018.

\bibitem{farhadi2009describing}
A.~Farhadi, I.~Endres, D.~Hoiem, and D.~Forsyth, ``Describing objects by their
  attributes,'' in {\em CVPR}, pp.~1778--1785, 2009.

\bibitem{lampert2009learning}
C.~H. Lampert, H.~Nickisch, and S.~Harmeling, ``Learning to detect unseen
  object classes by between-class attribute transfer,'' in {\em CVPR},
  pp.~951--958, 2009.

\bibitem{lampert2014attribute}
C.~H. Lampert, H.~Nickisch, and S.~Harmeling, ``Attribute-based classification
  for zero-shot visual object categorization,'' {\em TPAMI}, vol.~36, no.~3,
  pp.~453--465, 2014.

\bibitem{Akata2013}
Z.~Akata, F.~Perronnin, Z.~Harchaoui, and C.~Schmid, ``Label-embedding for
  attribute-based classification,'' in {\em CVPR}, 2013.

\bibitem{Romera-Paredes2015}
B.~Romera-Paredes and P.~H.~S. Torr, ``An embarrassingly simple approach to
  zero-shot learning,'' in {\em ICML}, pp.~2152--2161, 2015.

\bibitem{akata2015evaluation}
Z.~Akata, S.~Reed, D.~Walter, H.~Lee, and B.~Schiele, ``Evaluation of output
  embeddings for fine-grained image classification,'' in {\em CVPR},
  pp.~2927--2936, 2015.

\bibitem{akata2016label}
Z.~Akata, F.~Perronnin, Z.~Harchaoui, and C.~Schmid, ``Label-embedding for
  image classification,'' {\em TPAMI}, vol.~38, no.~7, pp.~1425--1438, 2016.

\bibitem{Socher2013}
R.~Socher, M.~Ganjoo, C.~D. Manning, and A.~Y. Ng, ``Zero-shot learning through
  cross-modal transfer,'' in {\em NIPS}, pp.~935--943, 2013.

\bibitem{xian2016latent}
Y.~Xian, Z.~Akata, G.~Sharma, Q.~Nguyen, M.~Hein, and B.~Schiele, ``Latent
  embeddings for zero-shot classification,'' in {\em CVPR}, pp.~69--77, 2016.

\bibitem{Zhang2017}
L.~Zhang, T.~Xiang, and S.~Gong, ``Learning a deep embedding model for
  zero-shot learning,'' in {\em CVPR}, 2017.

\bibitem{Norouzi2013}
M.~Norouzi, T.~Mikolov, S.~Bengio, Y.~Singer, J.~Shlens, A.~Frome, G.~S.
  Corrado, and J.~Dean, ``Zero-shot learning by convex combination of semantic
  embeddings,'' {\em arXiv preprint arXiv:1312.5650}, 2013.

\bibitem{zhang2015}
Z.~Zhang and V.~Saligrama, ``Zero-shot learning via semantic similarity
  embedding,'' in {\em ICCV}, pp.~4166--4174, 2015.

\bibitem{changpinyo2016synthesized}
S.~Changpinyo, W.-L. Chao, B.~Gong, and F.~Sha, ``Synthesized classifiers for
  zero-shot learning,'' in {\em CVPR}, pp.~5327--5336, 2016.

\bibitem{kipf2016semi}
T.~N. Kipf and M.~Welling, ``Semi-supervised classification with graph
  convolutional networks,'' in {\em ICLR}, 2016.

\bibitem{goodfellow2014generative}
I.~Goodfellow, J.~Pouget-Abadie, M.~Mirza, B.~Xu, D.~Warde-Farley, S.~Ozair,
  A.~Courville, and Y.~Bengio, ``Generative adversarial nets,'' in {\em NIPS},
  pp.~2672--2680, 2014.

\bibitem{Zhu_2018_CVPR}
Y.~Zhu, M.~Elhoseiny, B.~Liu, X.~Peng, and A.~Elgammal, ``A generative
  adversarial approach for zero-shot learning from noisy texts,'' in {\em
  CVPR}, 2018.

\bibitem{Chen_2018_CVPR}
L.~Chen, H.~Zhang, J.~Xiao, W.~Liu, and S.-F. Chang, ``Zero-shot visual
  recognition using semantics-preserving adversarial embedding networks,'' in
  {\em CVPR}, 2018.

\bibitem{Verma_2018_CVPR}
V.~Kumar~Verma, G.~Arora, A.~Mishra, and P.~Rai, ``Generalized zero-shot
  learning via synthesized examples,'' in {\em CVPR}, 2018.

\bibitem{Xian_2018_CVPR}
Y.~Xian, T.~Lorenz, B.~Schiele, and Z.~Akata, ``Feature generating networks for
  zero-shot learning,'' in {\em CVPR}, 2018.

\bibitem{WelinderEtal2010}
P.~Welinder, S.~Branson, T.~Mita, C.~Wah, F.~Schroff, S.~Belongie, and
  P.~Perona, ``{Caltech-UCSD Birds 200},'' Tech. Rep. CNS-TR-2010-001,
  California Institute of Technology, 2010.

\bibitem{Scheirer_2013_TPAMI}
W.~J. Scheirer, A.~Rocha, A.~Sapkota, and T.~E. Boult, ``Towards open set
  recognition,'' {\em TPAMI}, vol.~35, 2013.

\bibitem{Bendale2015}
A.~Bendale and T.~Boult, ``Towards open world recognition,'' in {\em CVPR},
  pp.~1893--1902, 2015.

\bibitem{Rudd2018}
E.~M. Rudd, L.~P. Jain, W.~J. Scheirer, and T.~E. Boult, ``The extreme value
  machine,'' {\em TPAMI}, vol.~40, no.~3, pp.~762--768, 2018.

\bibitem{Lafferty2001}
J.~Lafferty, A.~McCallum, and F.~Pereira, ``Conditional random fields:
  Probabilistic models for segmenting and labeling sequence data,'' in {\em
  ICML}, pp.~282--289, 2001.

\bibitem{He2016}
K.~He, X.~Zhang, S.~Ren, and J.~Sun, ``Deep residual learning for image
  recognition,'' in {\em CVPR}, pp.~770--778, 2016.

\bibitem{Bojanowski2017}
P.~Bojanowski, E.~Grave, A.~Joulin, and T.~Mikolov, ``Enriching word vectors
  with subword information,'' in {\em TACL}, vol.~5, pp.~135--146, 2017.

\bibitem{Mikolov2013a}
T.~Mikolov, K.~Chen, G.~Corrado, and J.~Dean, ``Efficient estimation of word
  representations in vector space,'' {\em arXiv preprint arXiv:1301.378}, 2013.

\bibitem{Mikolov2013b}
T.~Mikolov, I.~Sutskever, K.~Chen, G.~Corrado, and J.~Dean, ``Distributed
  representations of words and phrases and their compositionality,'' in {\em
  NIPS}, pp.~3111--3119, 2013.

\bibitem{Pennington2014}
J.~Pennington, R.~Socher, and C.~D. Manning, ``{GloVe: Global Vectors for Word
  Representation},'' in {\em EMNLP}, pp.~1532--1543, 2014.

\bibitem{imagenet}
O.~Russakovsky, J.~Deng, H.~Su, J.~Krause, S.~Satheesh, S.~Ma, Z.~Huang,
  A.~Karpathy, A.~Khosla, M.~Bernstein, {\em et~al.}, ``Imagenet: large scale
  visual recognition challenge,'' {\em International Journal of Computer
  Vision}, vol.~115, no.~3, pp.~211--252, 2015.

\bibitem{wordnet}
G.~A. Miller, ``Wordnet: a lexical database for {E}nglish,'' {\em
  Communications of the ACM}, vol.~38, no.~11, pp.~39--41, 1995.

\bibitem{Edmonds1967}
J.~Edmonds, ``{Optimum Branchings},'' {\em Journal of Research of the National
  Bureau of Standards}, vol.~71B, no.~4, p.~233, 1967.

\end{thebibliography}
}

\end{document}